\documentclass{article}

\usepackage[english]{babel}
\usepackage[letterpaper,top=2cm,bottom=2cm,left=3cm,right=3cm,marginparwidth=1.75cm]{geometry}
\usepackage[T1]{fontenc}
\usepackage{amsmath}
\usepackage{graphicx}
\usepackage[colorlinks=true, allcolors=blue]{hyperref}
\usepackage{float}
\usepackage{tabularray}
\usepackage{listings}
\usepackage{xcolor}
\usepackage{caption}
\usepackage{subcaption}

\title{THaLLE-ThaiLLM: Domain-Specialized Small LLMs for Finance and Thai -- Technical Report}
\author{NLP-Voice Research Lab, KBTG Labs,\\
\textit{KASIKORN Business---Technology Group}\\}
\date{January 1, 2026}

\begin{document}
\maketitle

\begin{abstract}
  Large Language Models (LLMs) have demonstrated significant potential across various domains, particularly in banking and finance, where they can automate complex tasks and enhance decision-making at scale.
  Due to privacy, security, and regulatory concerns, organizations often prefer on-premise deployment of LLMs.
  The ThaiLLM initiative aims to enhance Thai language capabilities in open-LLMs, enabling Thai industry to leverage advanced language models.
  However, organizations often face a trade-off between deploying multiple specialized models versus the prohibitive expense of training a single multi-capability model. 
  To address this, we explore model merging as a resource-efficient alternative for developing high-performance, multi-capability LLMs.
  We present results from two key experiments: first, merging Qwen-8B with ThaiLLM-8B demonstrates how ThaiLLM-8B enhances Thai general capabilities, showing an uplift of M3 and M6 O-NET exams over the general instruction-following Qwen-8B.
  Second, we merge Qwen-8B with both ThaiLLM-8B and THaLLE-CFA-8B.
  This combination results in further improvements in performance across both general and financial domains, by demonstrating an uplift in both M3 and M6 O-NET, Flare-CFA, and Thai-IC benchmarks.
  The report showcases the viability of model merging for efficiently creating multi-capability LLMs.
\end{abstract}

\section{Introduction}

  Large Language Models (LLMs) have seen rapid advancements in recent years, with both closed-source models (e.g., GPT-4 \cite{openai2024gpt4}, Gemini \cite{geminiteam2025geminifamilyhighlycapable}) and open-source models (e.g., LLaMA \cite{grattafiori2024llama3herdmodels}, Qwen \cite{deepyang2025qwen3technicalreport}) pushing the boundaries of natural language understanding and generation.
  In banking and finance sector, open-source LLMs offer significant advantages, particularly regarding privacy, data regulation compliance, and cost efficiency. \cite{labs2024thalletexthyperlocallyaugmented}
  By deploying models on-premise, financial institutions can ensure sensitive customer data remains secure and compliant with strict regulatory frameworks, while also avoiding the unpredictable costs and availability associated with API-based services.

  Current LLMs face significant challenges and limitations, particularly in the context of Thai language and specialized financial domains.
  Most global models are predominantly trained on English \cite{grattafiori2024llama3herdmodels, jiang2023mistral7b, openai2025gptoss120bgptoss20bmodel} and Chinese \cite{deepyang2025qwen3technicalreport, DeepSeekAI2024DeepSeekV3TR, Bai2025KimiKO} data, resulting in suboptimal performance when processing Thai text or understanding local financial nuances.
  ThaiLLM initiative aims to bridge this gap by enhancing Thai language capabilities in open-source LLMs, effectively enabling Thai industry to leverage advanced language models.

  Open-source LLMs allow organizations to fine-tune models on their own data to further improve capabilities in their domain of interest.
  However, leveraging rapid advancement in open-source LLMs is challenging due to high cost and complexity of retraining specialized models on newly released open-source LLMs. Maintaining a dedicated model for every specialized task is becoming unsustainable as the range of target applications grows. Furthermore, smaller specialized models often have limited capabilities, making it challenging to select the right set of specialized tasks to train into a single model.
  The decision space for task selection is exponential in nature ($2^T$ where $T$ is the number of tasks).

  Building upon the financial domain adaptation technique established in THaLLE-CFA \cite{labs2024thalletexthyperlocallyaugmented}, we introduce ThaiLLM initiative. To balance performance with computational efficiency, we employ model merging as a lightweight composition framework \cite{Goddard2024ArceesMA}. This approach facilitates the independent training of task-specific models, which are subsequently integrated into a singular, multi-capability architecture through model merging.

\section{Background}

  This section presents an overview of the evaluation benchmarks used in this study, including academic and financial related exams, along with the methodologies. This is to  enhance model performance, specifically Low-Rank Adaptation (LoRA) and model merging for LLMs.

  \subsection{Ordinary National Educational Test}
    The Ordinary National Educational Test (O-NET) is a nationwide standardized examination administered in Thailand to assess students' academic performance at critical educational milestones. 
    The exam evaluates student performance across five core subjects: Thai Language, Science, Mathematics, Social Studies, and English.
    The O-NET is administered at the lower secondary (M3, Grade 9) and upper secondary (M6, Grade 12) levels.
    This multi-level assessment provides a comprehensive framework for evaluating Thai language comprehension and reasoning capabilities across diverse academic domains.

  \subsection{Flare CFA}
    Flare Chartered Financial Analyst (CFA) is a specialized dataset designed to evaluate financial domain comprehension, with particular emphasis on topics encompassed within CFA examination curriculum. 
    This benchmark is particularly relevant for assessing language models' capacity to comprehend investment principles, financial analysis methodologies, and professional standards requisite for CFA certification, thereby serving as a critical evaluation metric for models deployed in the financial sector.

  \subsection{Investment Consultant Exam}
    The Investment Consultant (IC) Exam is a professional licensing issued by the Stock Exchange of Thailand (SET) to evaluate financial expertise and regulatory knowledge.
    The exam is a mandatory certification for individuals providing professional investment advice across a wide range of financial products.
    The examination comprises three components: P1 (fundamental knowledge), P2 (investment products), and P3 (investment planning and portfolio management). 
    This exam provides a framework for evaluating comprehension of Thai financial regulations, investment principles, and practical advisory competencies.

  \subsection{Low-Rank Adaptation of Large Language Models}
    Low-Rank Adaptation (LoRA) is a resource-efficient fine-tuning methodology designed to adapt LLMs to domain-specific tasks without modifying the entirety of model parameters \cite{hu2022lora}. 
    The fundamental principle of LoRA involves decomposing weight updates into low-rank matrices, thereby substantially reducing the number of trainable parameters while preserving model performance. 
    This methodology confers several advantages, including improved memory efficiency, faster training procedures, and reduced overfitting.

    Despite its demonstrated efficiency and efficacy, models fine-tuned with LoRA may experience limited learning capability due to its low rank nature.
    These limitations can be overcome by performing multiple sequential LoRA fine-tuning steps to achieve higher-rank updates \cite{lialin2023relora}.

  \subsection{Model Weight Merging}
    Model weight merging refers to a class of techniques that combine multiple pre-trained or fine-tuned checkpoints into a single model intended to inherit useful behaviors from each constituent model.
    A convenient way to view a checkpoint is as a base set of parameters together with an accumulated update induced by a particular training objective.
    Under this perspective, $W_{\text{merge}}$ is a linear interpolation in parameter space over the constituent checkpoints, and can be rewritten as a weighted update applied to the shared base parameters.
    
    \begin{equation}
    \lambda_i = \frac{w_i}{\sum_{j=1}^{n} w_j},
    \qquad
    W_{\text{merge}} = \sum_{i=1}^{n} \lambda_i W_i
    = W_{\text{base}} + \sum_{i=1}^{n} \lambda_i \Delta W_i
    \end{equation}
    where $W_{\text{merge}}$ denotes the merged model parameters, $W_{\text{base}}$ denotes the common base checkpoint, $\lambda_i$ is the normalized merging coefficient for model (or task) $i$, and $\Delta W_i = W_i - W_{\text{base}}$ represents the task-specific update relative to the base.

\section{Experiment}
  In this section, we cover the evaluation benchmarks and the model merging experiments conducted.

  \subsection{Evaluation}
    To comprehensively assess model performance, we conducted evaluations\footnote{the evaluations were conducted using \href{https://github.com/vllm-project/vllm}{vLLM}} across five distinct domains: academic, financial, safety, and prompt adherence.
    Since Qwen3-8B \cite{yang2025qwen3} facilitates transitions between reasoning and non-reasoning modes within a single LLM, all experimental model weights were evaluated under both configurations.
    Prompts used for evaluation are outlined in Appendix~\ref{apx:eval_prompt}.

    \subsubsection{Ordinary National Educational Test}
      The publicly available OpenThaiEval benchmark contains the O-NET dataset\footnote{\href{https://huggingface.co/datasets/iapp/openthaieval}{iapp/openthaieval}}.
      Given that ThaiLLM was developed to enhance performance on Thai language tasks, O-NET was selected as a benchmark to evaluate the model's ability to answer examination questions in Thai. 
      Since the evaluated LLMs are not multimodal, questions containing images and visual understanding were removed from the evaluation set.
      The number of questions for M3 and M6 O-NET exams are listed in Table~\ref{tab:onet_data}.

      \begin{table}[H]
        \centering
        \begin{tblr}{
          colspec = {c l c c},
          cell{2}{1} = {r=5}{c},
          cell{7}{1} = {r=5}{c},
          row{1} = {font=\bfseries},
          hline{1,2,7,12} = {-}{},
        }
          Level & Subject & Total & Remaining \\
          M3 & Thai Language & 29 & 29 \\
            & Social Studies & 20 & 20 \\
            & Mathematics & 20 & 20 \\
            & Science & 41 & 26 \\
            & English & 32 & 30 \\
          M6 & Thai Language & 65 & 65 \\
            & Social Studies & 60 & 60 \\
            & Mathematics & 25 & 19 \\
            & Science & 45 & 28 \\
            & English & 60 & 52 \\
        \end{tblr}
        \caption{Number of O-NET questions before and after filtering multimodal content.}
        \label{tab:onet_data}
      \end{table}

    \subsubsection{Flare CFA}
      Flare CFA is a publicly available benchmark based on the CFA exam\footnote{\href{https://huggingface.co/datasets/TheFinAI/flare-cfa}{TheFinAI/flare-cfa}}.
      This dataset comprises 1,032 questions covering CFA exam levels I and II.

    \subsubsection{Thai Investment Consultant}
      The publicly available OpenThaiEval benchmark contains the Investment Consultant (IC) license examination\footnote{\href{https://huggingface.co/datasets/iapp/openthaieval}{iapp/openthaieval}}.
      The dataset comprises 25 questions drawn from all three levels (P1, P2, and P3).

    \subsubsection{ThaiSafetyBench}
      ThaiSafetyBench is a publicly available benchmark for evaluating model safety\footnote{\href{https://huggingface.co/datasets/anonymoussssssss/ThaiSafetyBench}{anonymoussssssss/ThaiSafetyBench}}.
      It consists of Thai-language prompts with malicious and policy-violating instructions across multiple categories.
      The dataset includes both translated malicious prompts and prompts specifically crafted to reflect Thai cultural contexts, enabling wider assessment coverage of prompt injection attacks.
      The dataset contains a total of 1,954 prompts.

    \subsubsection{Thai-Output Consistency Test}
      We utilize a Thai-adapted version of the IFEval dataset\footnote{\href{https://huggingface.co/datasets/scb10x/ifeval-th}{scb10x/ifeval-th}}, which consists of 500 prompts tailored to Thai linguistic contexts. This benchmark serves to evaluate the model's ability to adhere strictly to language-specific constraints, specifically measuring its consistency in generating Thai-language outputs.
      
  \subsection{Model Merging}

    We utilized MergeKit \cite{Goddard2024ArceesMA} to merge multiple models into a single multi-capability model.
    Henceforth, we will refer to models trained on top of a common ancestor model as ``base models'' since they are a candidate for merging.
    The following list are the base models that were independently developed and utilized in our model merging experiments:

    \begin{enumerate}
        \item Qwen3-8B: An instruction-following model derived from Qwen3-8B-Base developed by Qwen team \cite{yang2025qwen3}.
        The model undergoes multiple stages of training to enable large language models to accurately follow human instructions and perform effectively in real-world applications.
        In addition, a dedicated safety layer is incorporated through fine-tuning to ensure responsible and reliable behavior.
        \item ThaiLLM-8B: A foundation model variant derived from Qwen3-8B-Base, developed by ThaiLLM Project.
        The model was enhanced through Continued Pre-Training (CPT) on 63 billion Thai-language text tokens to strengthen Thai-specific linguistic and domain knowledge.
        \item THaLLE-Finance-8B: Our supervised fine-tuned variant of Qwen3-8B that enhances financial domain knowledge through multiple rounds of low-rank Supervised Fine-Tuning (SFT) \cite{lialin2023relora}.
    \end{enumerate}

    We conducted two model merging experiments using linear merging.
    Linear merging is a straightforward yet effective approach to model combination, wherein each parameter is a weighted average of the respective parameters of the checkpoints.

    \begin{enumerate}
        \item ThaiLLM-8B-Instruct: Linear merging Qwen3-8B with ThaiLLM-8B with equal weight to create a model that can follow instructions and perform well on general Thai language tasks.
        \item THaLLE-0.2-ThaiLLM-8B-fa: Linear merging Qwen3-8B, ThaiLLM-8B, and THaLLE-Finance-8B with equal weight.
    \end{enumerate}

    The first experiment aims to demonstrate transferring instruction-following capabilities to ThaiLLM-8B, which is a pretrained foundation model.
    The second experiment aims to demonstrate the effectiveness of merging multiple models with differing capabilities to create a multi-capability model that performs well on both general and financial domains.
    The summary details for each base model and merged model are shown in Tables~\ref{tab:base_model_info},~\ref{tab:merge_model_info} respectively.
    Merge configurations are outlined in Appendix~\ref{apx:merge_cfg}

    \begin{table}[H]
      \centering
      \begin{tblr}{
        cells = {c},
        hline{1-2,5} = {-}{},
      }
        \textbf{Model}        & \textbf{Base}  & \textbf{Training Techniques} & \textbf{Objective}          \\
            Qwen3-8B          & Qwen3-8B-Base  & Multiples                        & Instruction Following       \\
            ThaiLLM-8B        & Qwen3-8B-Base  & CPT                              & Thai Language Understanding \\
            THaLLE-Finance-8B & Qwen3-8B       & SFT                              & Financial Knowledge         \\
      \end{tblr}
      \caption{Model training techniques and objective}
      \label{tab:base_model_info}
    \end{table}

    \begin{table}[H]
      \centering
      \begin{tblr}{
        cells = {c},
        hline{1-2,4} = {-}{},
      }
        \textbf{Merged Model}    & \textbf{Components}  \\
        ThaiLLM-8B-Instruct   & Qwen3-8B, ThaiLLM-8B              \\
        THaLLE-0.2-ThaiLLM-8B-fa   & Qwen3-8B, ThaiLLM-8B, THaLLE-Finance-8B              \\
      \end{tblr}
      \caption{Merged models and corresponding components}
      \label{tab:merge_model_info}
    \end{table}

\section{Results}\label{sec:results}
  The results of our evaluations are presented in this section, encompassing both general and financial domain performance, model safety assessments, and prompt adherence.

  \subsection{General Domain and Financial Domain}

    The evaluation results for base models and merged models on general-domain (O-NET) and financial-domain (CFA, IC) tasks are summarized in Table~\ref{tab:eval}.

    Our first merged model, ThaiLLM-8B-Instruct, demonstrates enhanced performance in both general domain and financial domain in both reasoning and non-reasoning modes over the base instruction following Qwen3-8B.
    The model attains scores of 0.707 on M3 and 0.623 on M6, surpassing Qwen3-8B across all evaluated benchmarks.

    Our second merged model, THaLLE-0.2-ThaiLLM-8B-fa, substantiates further performance uplift in both general domain and financial domain in both reasoning and non-reasoning modes, with an exception of the M6 O-NET in non-reasoning mode.
    In particular, THaLLE-0.2-ThaiLLM-8B-fa achieves the highest scores on the CFA and Thai-IC benchmarks, with scores of 0.771 and 0.720, respectively.
    Overall, THaLLE-0.2-ThaiLLM-8B-fa yields performance improvements of 12.6\% on O-NET, 5.7\% on the CFA benchmark, and 40\% on the Thai-IC exam over the base Qwen3-8B model, when reasoning mode is enabled.

    \begin{table}[ht]
    \centering
    \small
    \begin{tblr}{
      colspec = {l c c c c},
      row{1,2} = {font=\bfseries},
      cell{1}{1} = {r=2}{c},
      cell{1}{2} = {c=2}{c},
      cell{1}{4} = {r=2}{c},
      cell{1}{5} = {r=2}{c},
      hline{1,3,7,11} = {-}{},
    }
    Model
      & O-NET
      & 
      & CFA
      & IC \\

      & M3
      & M6
      &  & \\

    \textit{Non-Reasoning} \\
    Qwen3-8B \cite{yang2025qwen3}
      & 0.660          & 0.545          & 0.753          & 0.640 \\
    ThaiLLM-8B-Instruct
      & 0.707          & \textbf{0.623} & 0.762          & \textbf{0.720} \\
    THaLLE-0.2-ThaiLLM-8B-fa
      & \textbf{0.725} & 0.572          & \textbf{0.771} & \textbf{0.720} \\

    \textit{Reasoning} \\
    Qwen3-8B
      & 0.706 & 0.590 & 0.806 & 0.600 \\
    ThaiLLM-8B-Instruct
      & 0.720 & 0.661 & 0.820 & 0.720 \\
    THaLLE-0.2-ThaiLLM-8B-fa
      & \textbf{0.779} & \textbf{0.678} & \textbf{0.852} & \textbf{0.840} \\
    \end{tblr}

    \caption{Evaluation results on general-domain (O-NET) and financial-domain (CFA, IC) tasks.}
    \label{tab:eval}
    \end{table}

  \subsection{Model Safety}

    The evaluation with and without explicit safety instructions on ThaiSafetyBench is reported in Table~\ref{tab:safety_eval}.
    In the absence of explicit safety instructions, Qwen model family, including our experimental merges, exhibits poor performance on this benchmark.
    However, the provision of safety instructions results in a significant improvement in safety performance for both ThaiLLM-8B-Instruct and THaLLE-0.2-ThaiLLM-8B-fa in non-reasoning mode.
 In particular, this safety boost is not observed in reasoning mode.
    We attribute this to the fact that our fine-tuning process did not include specific safety training and alignment for reasoning mechanisms.

    We recommend incorporating explicit safety instructions for the deployment of our models in consumer-facing applications to ensure robust handling of potentially harmful prompts.

    \begin{table}[!ht]
      \centering
      \begin{tblr}{
        colspec = {l c c},
        row{1,2} = {font=\bfseries},
        cell{1}{1} = {r=2}{c},
        cell{1}{2} = {c=2}{c},
        cell{3}{1} = {c=2}{l},
        cell{7}{1} = {c=2}{l},
        hline{1,3,7,11} = {-}{},
      }

        Model & ThaiSafetyBench \\
        & without safety inst. & with safety inst. \\

        \textit{Non-Reasoning} \\
        Qwen3-8B & 0.346 & 0.853 \\
        ThaiLLM-8B & 0.268 & 0.924 \\
        THaLLE-0.2-ThaiLLM-8B-fa & 0.300 & \textbf{0.947} \\

        \textit{Reasoning} \\
        Qwen3-8B & 0.197 & \textbf{0.810} \\
        ThaiLLM-8B & 0.274 & 0.753 \\
        THaLLE-0.2-ThaiLLM-8B-fa & 0.254 & 0.794 \\

      \end{tblr}
      \caption{Model refusal rates for potentially harmful requests (higher is better)}
      \label{tab:safety_eval}
    \end{table}

  \subsection{Thai Output Consistency}

    The results for the IFEval-TH benchmark are presented in Table~\ref{tab:ifeval}.
    Both model merges achieve superior performance on Thai output consistency tests.
    ThaiLLM-8B-Instruct model exhibits competitive performance relative to THaLLE-0.2-ThaiLLM-8B-fa, with scores of 0.994 and 0.982 in non-reasoning mode, and 0.964 and 0.976 in reasoning mode, respectively.

    This finding highlights the effectiveness of continued pre-training with Thai-language context in improving the model's ability to understand and consistently generate Thai outputs, compared with the base model (Qwen3-8B).

    \begin{table}[!ht]
      \centering
      \begin{tblr}{
        colspec = {l c},
        row{1} = {font=\bfseries},
        cell{2}{1} = {c=2}{l},
        cell{6}{1} = {c=2}{l},
        hline{1-2,6,10} = {-}{},
      }

        Model & IFEval-TH \\

        \textit{Non-Reasoning} & \\
        Qwen3-8B & 0.944 \\
        ThaiLLM-8B & \textbf{0.994} \\
        THaLLE-0.2-ThaiLLM-8B-fa & 0.982 \\

        \textit{Reasoning} & \\
        Qwen3-8B & 0.926 \\
        ThaiLLM-8B & 0.964 \\
        THaLLE-0.2-ThaiLLM-8B-fa & \textbf{0.976} \\

      \end{tblr}
      \caption{Model performance on Thai output consistency using IFEval-TH}
      \label{tab:ifeval}
    \end{table}

\section{Conclusion}\label{sec:conclusion}

  In this study, we investigate model merging as a computationally efficient yet robust strategy for enhancing open-source large language models across specialized domains, specifically Thai language proficiency and financial expertise.
  Our experimental results demonstrate that this approach effectively facilitates performance gains in specific tasks (ThaiLLM-8B-Instruct), as well as a unified model with complementary capabilities (THaLLE-0.2-ThaiLLM-8B-fa).

  A key advantage of the proposed approach is its computational efficiency.
  Model merging allows checkpoints to be developed independently, forming modular building blocks that can be combined as needed.
  The merging process requires minimal resources and can be executed on hardware without GPUs, making it particularly suitable for early-stage experimentation, rapid prototyping, and domain-specific adaptation of large language models under resource constraints.

  Although the resulting merged models exhibit strong performance across evaluated benchmarks, we believe that there remains potential for further enhancement through additional fine-tuning or from more sophisticated merging techniques.
  Such extensions are beyond the scope of this work and are left for future investigation.
  We hope that this study provides useful empirical evidence for researchers and practitioners, highlighting model merging as a practical, low-resource alternative for adapting large language models and contributing to the continued advancement of open-source language model development.

\section*{Contributions and Acknowledgments}

  We extend our gratitude to the executive team for their leadership and support.\newline
  \newline
  \textit{Core Contributors:} Anuruth Lertpiya, Danupat Khamnuansin, Kantapong Sucharitpongpan, Pornchanan Balee
  \newline
  \textit{Executive Leadership\footnote{\label{footnote:name}By alphabetical order}:} Monchai Lertsutthiwong, Tawunrat Chalothorn, Thadpong Pongthawornkamol

\newpage
\bibliographystyle{unsrt}
\bibliography{ref}

@misc{openai2024gpt4,
      title={GPT-4 Technical Report}, 
      author={OpenAI and others},
      year={2024},
      eprint={2303.08774},
      archivePrefix={arXiv},
      primaryClass={cs.CL}
}

@misc{geminiteam2025geminifamilyhighlycapable,
      title={Gemini: A Family of Highly Capable Multimodal Models}, 
      author={Gemini Team and others},
      year={2025},
      eprint={2312.11805},
      archivePrefix={arXiv},
      primaryClass={cs.CL},
      url={https://arxiv.org/abs/2312.11805}, 
}

@misc{grattafiori2024llama3herdmodels,
      title={The Llama 3 Herd of Models}, 
      author={Llama Team and others},
      year={2024},
      eprint={2407.21783},
      archivePrefix={arXiv},
      primaryClass={cs.AI},
      url={https://arxiv.org/abs/2407.21783}, 
}

@misc{deepyang2025qwen3technicalreport,
      title={Qwen3 Technical Report}, 
      author={An Yang and Anfeng Li and Baosong Yang and Beichen Zhang and Binyuan Hui and Bo Zheng and Bowen Yu and Chang Gao and Chengen Huang and Chenxu Lv and Chujie Zheng and Dayiheng Liu and Fan Zhou and Fei Huang and Feng Hu and Hao Ge and Haoran Wei and Huan Lin and Jialong Tang and Jian Yang and Jianhong Tu and Jianwei Zhang and Jianxin Yang and Jiaxi Yang and Jing Zhou and Jingren Zhou and Junyang Lin and Kai Dang and Keqin Bao and Kexin Yang and Le Yu and Lianghao Deng and Mei Li and Mingfeng Xue and Mingze Li and Pei Zhang and Peng Wang and Qin Zhu and Rui Men and Ruize Gao and Shixuan Liu and Shuang Luo and Tianhao Li and Tianyi Tang and Wenbiao Yin and Xingzhang Ren and Xinyu Wang and Xinyu Zhang and Xuancheng Ren and Yang Fan and Yang Su and Yichang Zhang and Yinger Zhang and Yu Wan and Yuqiong Liu and Zekun Wang and Zeyu Cui and Zhenru Zhang and Zhipeng Zhou and Zihan Qiu},
      year={2025},
      eprint={2505.09388},
      archivePrefix={arXiv},
      primaryClass={cs.CL},
      url={https://arxiv.org/abs/2505.09388}, 
}

@misc{labs2024thalletexthyperlocallyaugmented,
      title={THaLLE: Text Hyperlocally Augmented Large Language Extension -- Technical Report}, 
      author={KBTG Labs and Danupat Khamnuansin and Atthakorn Petchsod and Anuruth Lertpiya and Pornchanan Balee and Thanawat Lodkaew and Tawunrat Chalothorn and Thadpong Pongthawornkamol and Monchai Lertsutthiwong},
      year={2024},
      eprint={2406.07505},
      archivePrefix={arXiv},
      primaryClass={cs.CL},
      url={https://arxiv.org/abs/2406.07505}, 
}

@misc{jiang2023mistral7b,
      title={Mistral 7B}, 
      author={Albert Q. Jiang and Alexandre Sablayrolles and Arthur Mensch and Chris Bamford and Devendra Singh Chaplot and Diego de las Casas and Florian Bressand and Gianna Lengyel and Guillaume Lample and Lucile Saulnier and Lélio Renard Lavaud and Marie-Anne Lachaux and Pierre Stock and Teven Le Scao and Thibaut Lavril and Thomas Wang and Timothée Lacroix and William El Sayed},
      year={2023},
      eprint={2310.06825},
      archivePrefix={arXiv},
      primaryClass={cs.CL},
      url={https://arxiv.org/abs/2310.06825}, 
}

@misc{openai2025gptoss120bgptoss20bmodel,
      title={gpt-oss-120b \& gpt-oss-20b Model Card}, 
      author={OpenAI},
      year={2025},
      eprint={2508.10925},
      archivePrefix={arXiv},
      primaryClass={cs.CL},
      url={https://arxiv.org/abs/2508.10925}, 
}

@inproceedings{Goddard2024ArceesMA,
  title={Arcee’s MergeKit: A Toolkit for Merging Large Language Models},
  author={Charles Goddard and Shamane Siriwardhana and Malikeh Ehghaghi and Luke Meyers and Vladimir Karpukhin and Brian Benedict and Mark McQuade and Jacob Solawetz},
  booktitle={Conference on Empirical Methods in Natural Language Processing},
  year={2024},
  url={https://api.semanticscholar.org/CorpusID:268537132}
}

@article{Bai2025KimiKO,
  title={Kimi K2: Open Agentic Intelligence},
  author={Kimi Team and others},
  journal={ArXiv},
  year={2025},
  volume={abs/2507.20534},
  url={https://api.semanticscholar.org/CorpusID:280323540}
}

@article{DeepSeekAI2024DeepSeekV3TR,
  title={DeepSeek-V3 Technical Report},
  author={DeepSeek-AI and others},
  journal={ArXiv},
  year={2024},
  volume={abs/2412.19437},
  url={https://api.semanticscholar.org/CorpusID:275118643}
}

@misc{lialin2023relora,
      title={ReLoRA: High-Rank Training Through Low-Rank Updates}, 
      author={Vladislav Lialin and Namrata Shivagunde and Sherin Muckatira and Anna Rumshisky},
      year={2023},
      eprint={2307.05695},
      archivePrefix={arXiv},
      primaryClass={cs.CL},
      url={https://arxiv.org/abs/2307.05695}, 
}

@misc{yang2025qwen3,
      title={Qwen3 Technical Report}, 
      author={An Yang and Anfeng Li and Baosong Yang and Beichen Zhang and Binyuan Hui and Bo Zheng and Bowen Yu and Chang Gao and Chengen Huang and Chenxu Lv and Chujie Zheng and Dayiheng Liu and Fan Zhou and Fei Huang and Feng Hu and Hao Ge and Haoran Wei and Huan Lin and Jialong Tang and Jian Yang and Jianhong Tu and Jianwei Zhang and Jianxin Yang and Jiaxi Yang and Jing Zhou and Jingren Zhou and Junyang Lin and Kai Dang and Keqin Bao and Kexin Yang and Le Yu and Lianghao Deng and Mei Li and Mingfeng Xue and Mingze Li and Pei Zhang and Peng Wang and Qin Zhu and Rui Men and Ruize Gao and Shixuan Liu and Shuang Luo and Tianhao Li and Tianyi Tang and Wenbiao Yin and Xingzhang Ren and Xinyu Wang and Xinyu Zhang and Xuancheng Ren and Yang Fan and Yang Su and Yichang Zhang and Yinger Zhang and Yu Wan and Yuqiong Liu and Zekun Wang and Zeyu Cui and Zhenru Zhang and Zhipeng Zhou and Zihan Qiu},
      year={2025},
      eprint={2505.09388},
      archivePrefix={arXiv},
      primaryClass={cs.CL},
      url={https://arxiv.org/abs/2505.09388}, 
}

@inproceedings{
    hu2022lora,
    title={Lo{RA}: Low-Rank Adaptation of Large Language Models},
    author={Edward J Hu and yelong shen and Phillip Wallis and Zeyuan Allen-Zhu and Yuanzhi Li and Shean Wang and Lu Wang and Weizhu Chen},
    booktitle={International Conference on Learning Representations},
    year={2022},
    url={https://openreview.net/forum?id=nZeVKeeFYf9}
}

\newpage

\section*{Appendix}
\appendix

\section{Evaluation Prompts}\label{apx:eval_prompt}

\begin{table}[H]
    \centering
    \begin{tabular}{|p{3.25cm}|p{11.5cm}|}
        \hline
        \textbf{Task} & \textbf{Prompt} \\
        \hline
        CFA \newline (User prompt) & {``You are a CFA (chartered financial analyst) taking a test to evaluate your knowledge of finance. You will be given a question along with three possible answers. Indicate the correct answer (A, B, or C).''} \\
        \hline
        IC \newline (User prompt) & {``You are a Certified Thai Investment Consultant (IC) taking a test to evaluate your knowledge of finance. You will be given a question along with four possible answers. Indicate the correct answer (1, 2, 3, or 4).''} \\
        \hline
        ThaiSafetyBench \newline (Safety instructions as system prompt) & {``You are a strictly safety-aligned assistant.
        For any harmful, illegal, sensitive, biased, or ambiguous request, you must refuse to answer.
        Your response must be a brief and polite refusal only.
        Do not provide additional information, suggestions, explanations, or alternatives.''} \\
        \hline
    \end{tabular}
    \caption{Prompts used for evaluation}
    \label{tab:eval_prompt}
\end{table}

\section{Merge Configurations}
\label{apx:merge_cfg}

\begin{figure}[H]
\centering

\begin{minipage}[t]{0.48\linewidth}
\begin{lstlisting}
merge_method: linear
dtype: bfloat16
models:
  - model: Qwen/Qwen3-8B
    parameters:
      weight: 1.0
  - model: ThaiLLM/ThaiLLM-8B
    parameters:
      weight: 1.0
tokenizer:
  source: base
base_model: Qwen/Qwen3-8B
\end{lstlisting}
\centering
\captionof{figure}{ThaiLLM-8B-Instruct config}
\end{minipage}
\hfill
\begin{minipage}[t]{0.48\linewidth}
\begin{lstlisting}
merge_method: linear
dtype: bfloat16
models:
  - model: Qwen/Qwen3-8B
    parameters:
      weight: 1.0
  - model: ThaiLLM/ThaiLLM-8B
    parameters:
      weight: 1.0
  - model: THaLLE-Finance-8B
    parameters:
      weight: 1.0
tokenizer:
  source: base
base_model: Qwen/Qwen3-8B
\end{lstlisting}
\centering
\captionof{figure}{THaLLE-0.2-ThaiLLM-8B-fa config}
\end{minipage}
\end{figure}

\end{document}